\documentclass[11pt,a4paper]{article}
\usepackage[hyperref]{acl2020}
\usepackage{times}
\usepackage{latexsym}

\usepackage{algorithm}  
\usepackage{algorithmic}
\usepackage{multirow}
\usepackage{CJKutf8}
\usepackage{graphicx}
\usepackage{subcaption}
\usepackage{xcolor}
\usepackage{booktabs}
\usepackage{enumitem}
\usepackage{eqparbox}

\usepackage{amsthm,amsmath,amsfonts,bm,xspace}
\usepackage{bbm}
\usepackage{color}

\def\eg{{\em e.g.,}\xspace}
\def\ie{{\em i.e.,}\xspace}

\newcommand{\comment}[1]{}

\newcommand{\reza}[1]{{\textcolor{green}{#1}}}

\def\figref#1{Figure~\ref{#1}}
\def\Figref#1{Figure~\ref{#1}}

\def\tabref#1{Table~\ref{#1}}

\def\secref#1{section~\ref{#1}}

\def\eqref#1{(\ref{#1})}

\def\Algref#1{Algorithm~\ref{#1}}

\def\1{\bm{1}}

\def\vb{{\bm{b}}}

\def\vx{{\bm{x}}}
\def\vy{{\bm{y}}}
\def\vz{{\bm{z}}}

\DeclareMathAlphabet{\mathsfit}{\encodingdefault}{\sfdefault}{m}{sl}
\SetMathAlphabet{\mathsfit}{bold}{\encodingdefault}{\sfdefault}{bx}{n}

\newcommand{\one}[1]{\mathbbm{1}{[#1]}}

\usepackage{hyperref}
\usepackage{url}
\usepackage{microtype}

\aclfinalcopy 

\title{Dynamic Programming Encoding for\\Subword Segmentation in Neural Machine Translation}

\author{Xuanli He \\
  Monash University \\
  \\\And
  Gholamreza Haffari \\
  Monash University \\
  \texttt{\{xuanli.he1,gholamreza.haffari\}@monash.edu}~~~~~\texttt{mnorouzi@google.com}\\\And
  Mohammad Norouzi \\
  Google Research \\
  \\}

\date{}

\begin{document}
\maketitle
\begin{abstract}
This paper introduces Dynamic Programming Encoding (DPE),
a new segmentation algorithm for tokenizing sentences into subword units.
We view the subword segmentation of an output sentence as a latent
variable that should be marginalized for learning and
inference.
A mixed character-subword transformer is proposed,
which enables exact log marginal likelihood estimation and exact MAP
inference to find output segmentations with maximum posterior probability.
DPE uses such a mixed character-subword transformer as a
means of pre-processing parallel data to segment output sentences using
dynamic programming.
Empirical results on machine translation suggest that DPE
is effective for segmenting output sentences and
can be combined with BPE dropout for
stochastic segmentation of source sentences.
DPE achieves an average improvement of 0.9 BLEU over
BPE~\cite{sennrich2016neural} and an average improvement of 0.55 BLEU
over BPE dropout~\cite{provilkov2019bpe} on several WMT
datasets including English $\leftrightarrow$ (German, Romanian, Estonian, Finnish, Hungarian).

\end{abstract}

\section{Introduction}
The segmentation of rare words into subword
units~\citep{sennrich2016neural, wu2016google} has become a critical
component of neural machine translation~\citep{transformer17} and
natural language understanding~\citep{devlin2019bert}.  Subword units
enable {\em open vocabulary} text processing with a negligible
pre-processing cost and help maintain a desirable balance between the
vocabulary size and decoding speed. Since subword vocabularies are built
in an unsupervised manner~\citep{sennrich2016neural, wu2016google}, they 
can be easily adopted for any language.

Given a fixed vocabulary of subword units, rare words can be segmented
into a sequence of subword units in different ways.  For instance,
``un+conscious'' and ``uncon+scious'' are both suitable segmentations
for the word ``unconscious''. This paper studies the impact of subword
segmentation on neural machine translation, given a fixed subword vocabulary, and presents a new
algorithm called {\em Dynamic Programming Encoding (DPE)}.

We identify three families of subword segmentation algorithms in neural machine translation:
\begin{enumerate}[topsep=0pt, partopsep=0pt, leftmargin=15pt, parsep=0pt, itemsep=2pt]
    \item Greedy algorithms: \citet{wu2016google} segment words by recursively selecting
    the longest subword prefix. \citet{sennrich2016neural} recursively combine adjacent
    word fragments that co-occur most frequently, starting from characters.
    \item Stochastic algorithms~\citep{kudo2018subword,provilkov2019bpe}
    draw multiple segmentations for source and target sequences resorting to
    randomization to improve robustness and generalization of translation models.
    \item Dynamic programming algorithms, studied here,
    enable exact marginalization of subword segmentations for
    certain sequence models.
\end{enumerate}

We view the subword segmentation of an output sentence in machine translation as a latent
variable that should be marginalized to obtain the probability
of the output sentence given the input.
On the other hand,
the segmentation of source sentences can be thought of as input features
and can be randomized as a form of data augmentation to improve translation robustness
and generalization. Unlike previous work, we recommend using two distinct segmentation algorithms
for tokenizing source and target sentences:
stochastic segmentation for source and dynamic programming for target sentences.

We present a new family of mixed character-subword transformers, for which simple dynamic programming algorithms
exist for exact marginalization and MAP inference of subword segmentations.
The time complexity of the dynamic programming algorithms is $O(TV)$, where $T$ is the length of the target
sentence in characters, and $V$ is the size of the subword vocabulary. By comparison, even computing
the conditional probabilities of subword units in an autoregressive model requires $O(TV)$ to estimate
the normalizing constant of the categorical distributions. Thus, our dynamic programming algorithm does
not incur additional asymptotic costs.
We use a lightweight mixed character-subword transformer as a means of
pre-processing translation datasets to segment output sentences using DPE for MAP inference.


The performance of a standard subword transformer \citep{transformer17}
trained on WMT datasets tokenized using DPE is compared against Byte Pair Encoding (BPE)~\citep{sennrich2016neural}
and BPE dropout~\cite{provilkov2019bpe}. 
Empirical results on
English $\leftrightarrow$ (German, Romanian, Estonian, Finnish, Hungarian)
suggest that stochastic subword segmentation
is effective for tokenizing source sentences,
whereas deterministic DPE  is superior for segmenting target sentences.
DPE achieves an average improvement of 0.9 BLEU over
greedy BPE~\cite{sennrich2016neural} and an average improvement of 0.55 BLEU 
over stochastic BPE dropout~\cite{provilkov2019bpe}\footnote{code and corpora: https://github.com/xlhex/dpe}.

\section{Related Work}
Neural networks have revolutionized machine translation \citep{sutskever2014sequence,dima15,cho2014learning}.
Early neural machine translation (NMT) systems used words as the atomic element of sentences.
They used vocabularies with tens of thousands words, resulting in prohibitive training and inference complexity.
While learning can be sped up using sampling techniques~\citep{jean2015using}, 
word based NMT models have a difficult time handling rare words, especially in morphologically rich languages such
as Romanian, Estonian, and Finnish. The size of the word vocabulary should increase dramatically to capture the compositionality of morphemes
in such languages.

More recently, many NMT models have been developed based on characters and a combination of characters and words~\cite{ling2015finding, luong2016achieving, Vylomova_2017,lee2017fully, cherry2018revisiting}.
Fully character based models~\cite{lee2017fully, cherry2018revisiting} demonstrate a significant improvement over word based models on morphologically rich languages. Nevertheless, owing to the lack of morphological information, deeper models are often required to obtain a good translation quality. Moreover, elongated sequences brought by a character representation drastically increases the inference latency.

In order to maintain a good balance between the vocabulary size and decoding speed, subword units are introduced in NMT~\cite{sennrich2016neural,wu2016google}. These segmentation approaches are data-driven and unsupervised.
Therefore, with a negligible pre-processing overhead, subword models can be applied to any NLP task \citep{transformer17,devlin2019bert}.
Meanwhile, since subword vocabularies are generated based on word frequencies,
only the rare words are split into subword units and common words remain intact.

Previous work~\citep{chan2016latent,kudo2018subword} has explored the idea of
using stochastic subword segmentation with multiple subword candidates to approximate the log marginal likelihood.
\citet{kudo2018subword} observed marginal gains in translation quality at the cost of introducing additional hyper-parameters
and complex sampling procedures. We utilize BPE dropout~\citep{provilkov2019bpe}, a simple stochastic segmentation
algorithm for tokenizing source sentences.

Dynamic programming has been used to marginalize out latent segmentations for speech recognition \cite{wang2017sequence},
showing a consistent improvement over greedy segmentation methods. In addition, dynamic programming has been successfully 
applied to learning sequence models by optimizing edit distance~\cite{sabour2018optimal} and aligning source and target
sequences~\citep{chan2020imputer,saharia2020non}. We show the effectiveness of dynamic programming for segmenting 
output sentences in NMT using a mixed character-transformer in a pre-processing step.

Prior to our work, multiple NLP tasks, such as word segmentation~\cite{kawakami-etal-2019-learning}, NMT~\cite{kreutzer2018learning} and language modeling~\cite{grave-etal-2019-training}, have been benefiting from segmentation via neural language models. However, to the best of our knowledge, this work is the first call on the conditional target subword segmentation.

\section{Latent Subword Segmentation}
\label{sec:subseg}

\begin{figure}[t]
\small
\centering
\includegraphics[scale=0.29]{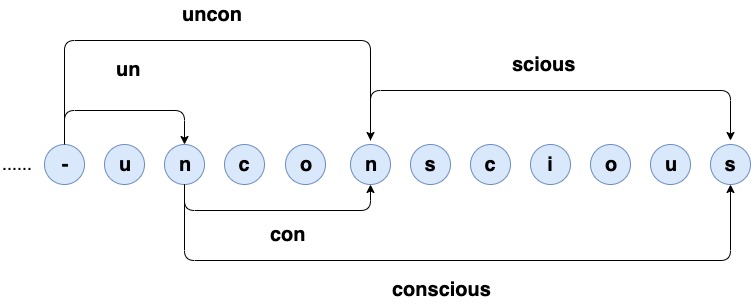}
\caption{\small An illustration of marginalizing subword segmentations of the word `unconscious'
}
\label{fig:ways}
\vspace{-.2cm}
\end{figure}

Let $\vx$ denote a source sentence and $\vy=(y_1, \ldots, y_T)$ denote a target sentence
comprising $T$ characters. The goal of machine translation is to learn a
conditional distribution $p(\vy \mid \vx)$ from a large corpus of source-target sentences.
State-of-the-art neural machine translation systems make use of a dictionary of subword units
to tokenize the target sentences in a more succinct way as a sequence of $M \le T$ subword units.
Given a subword vocabulary, there are multiple ways to segment a rare word into a sequence of subwords (see Figure \ref{fig:ways}).
The common practice in neural machine translation considers
subword segmentation as a pre-process and uses greedy algorithms
to segment each word across a translation corpus in a consistent way.
This paper 
aims to find optimal subword segmentations for the task of machine translation.

\renewcommand{\algorithmiccomment}[1]{\hfill\eqparbox{ }{$\triangleright$ #1}}

\begin{algorithm*}[t] 
\caption{Dynamic Programming (DP) for Exact Marginalization} 
\label{alg:dp} 
\begin{algorithmic}[1] 
\REQUIRE $\vy$ is a sequence of $T$ characters, $V$ is a subword vocabulary, $m$ is the maximum subword length
\ENSURE $\log p(\vy)$  marginalizing out different subword segmentations.
        \STATE {$\alpha_0 \leftarrow 0$}
         \FOR{$k=1$ \TO $T$}
         \STATE $\alpha_k \leftarrow \log \sum_{j=k-m}^{k-1} \one{\vy_{j,k} \in V}\exp\Big(\alpha_{j}+\log P_{\theta}(\vy_{j,k}|y_1,..,y_{j})\Big)$
            \ENDFOR
         \RETURN $\alpha_{T}$
          \COMMENT{ the marginal probability $\log p(\vy) = \log \sum_{\vz \in \mathcal{Z}_y}p(\vy, \vz)$}
\end{algorithmic}
\end{algorithm*}

Let $\vz = (z_1,..,z_{M+1})$ denote a sequence of character indices 
$0\!=\!z_1 < z_2 < \ldots < z_{M}<z_{M+1}\!=\!T$ in an ascending order, defining the boundary
of $M$ subword segments $\{\vy_{{z_i},{z_{i+1}}}\}_{i=1}^M$. Let
$
\vy_{a,b} ~\equiv~ [y_{a+1}, \ldots, y_b]~
$ 
denote a subword that spans the segment between $(a+1)^\text{th}$ and $b^\text{th}$ characters, including the boundary characters.
For example, given a subword dictionary \{`c', `a', `t', `at', `ca'\},
the word `cat' may be segmented using $\vz = (0,1,3)$ as (`c', `at'),
or using $\vz = (0,2,3)$ as (`ca', `t'),
or using $\vz = (0,1,2,3)$ as (`c', `a', `t').
The segmentation $\vz = (0,3)$ is not valid since `cat' does not appear in the subword vocabulary.

Autoregressive language models create a categorical distribution over the subword vocabulary at every subword position
and represent the log-probability of a subword sequence using chain rule,%
\begin{equation}
\begin{aligned}
    &\log p(\vy, \vz) ~=~\\
    &\sum\nolimits_{i=1}^{\lvert \vz \rvert} \log p (\vy_{{z_i},{z_{i+1}}} \mid \vy_{z_{1},{z_{2}}}, \ldots, \vy_{{z_{i-1}},{z_{i}}})~.
    \label{eq:individualz}
\end{aligned}
\end{equation}
Note that we suppress the dependence of $p$ on $\vx$ to reduce notational clutter. Most neural machine translation
approaches assume that $\vz$ is a deterministic function of $\vy$ and implicitly assume that $\log p(\vy, \vz) \approx \log p(\vy)$.

We consider a subword segmentation $\vz$ as a latent variable and let each value of $\vz \in \mathcal{Z}_y$, in
the set of segmentations compatible with $\vy$,
contribute its share to $p(\vy)$ according to $p(\vy) = \sum_\vz p(\vy, \vz)$,
\begin{equation}
\label{eq:jointz}
\begin{aligned}
  &\log p(\vy) ~=~\\ 
  &\log \sum_{\vz \in \mathcal{Z}_y} \exp \sum_{i=1}^{\lvert \vz \rvert}
  \log p(\vy_{{z_i},{z_{i+1}}} \mid \ldots, \vy_{{z_{i-1}},{z_{i}}})~. 
\end{aligned}
\end{equation}

Note that each particular subword segmentation $\vz \in \mathcal{Z}_y$ provides a lower bound on the log marginal likelihood $\log p(\vy) \ge \log p(\vy, \vz)$. Hence, optimizing \eqref{eq:individualz} for a greedily selected segmentation can be justified as a lower bound on \eqref{eq:jointz}.
That said, optimizing \eqref{eq:jointz} directly is more desirable.
Unfortunately, exact marginalization over all segmentations is computationally prohibitive in a combinatorially large space $\mathcal{Z}_y$,
especially because the probability of each subword depends on the segmentation of its conditioning context.
In the next section, we discuss a sequence model in which the segmentation of the conditioning context
does not influence the probability of the next subword. 
We describe an efficient Dynamic Programming algorithm to exactly marginalize out all possible subword segmentations in this model.

\section{A Mixed Character-Subword Transformer}

\begin{figure}[t]
\centering
\includegraphics[scale=0.45]{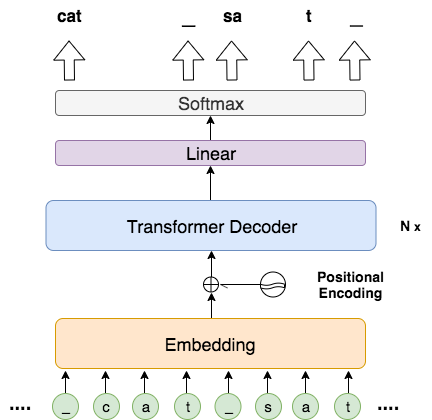}
\caption{An illustration of the mixed character-subword Transformer.
The input is a list of characters, whereas the output is a sequence of subwords.}
\label{fig:arch}
\vspace{-.2cm}
\end{figure}

We  propose a mixed character-subword transformer architecture, which enables one to marginalize out
latent subword segmentations exactly using dynamic programming (see Figure \ref{fig:arch}). 
Our key insight is to let the transformer architecture process the inputs and the conditioning context
based on characters to remain oblivious to the specific choice of subword segmentation in the conditioning context
and enable exact marginalization.
That said, the output of the transformer is based on subword units and at every position it creates
a categorical distribution over the subword vocabulary.
More precisely, when generating a subword $\vy_{{z_i},{z_{i+1}}}$, the model processes
the conditioning context $(y_{z_1}, \ldots, y_{z_i})$ based solely on characters using,
\begin{equation}
    \log p(\vy, \vz) = \sum\nolimits_{i=1}^{\lvert \vz \rvert}
    \log p(\vy_{{z_i},{z_{i+1}}} \mid y_{z_1},...,y_{z_i})~,
    \label{eq:jointz2}
\end{equation}
where  the dependence of $p$ on $\vx$ is suppressed to reduce notational clutter. 


Given a fixed subword vocabulary denoted $V$,
at every character position $t$ within $\vy$, the mixed character-subword model induces a distribution
over the next subword $w \in V$ based on,
\begin{equation}
\label{eqn:simp}
p(w \!\mid\! y_1,.., y_t) \!=\! \frac{\exp(f(y_1, .., y_t)^\top e(w))}{\sum_{w' \in V}\exp(f(y_1, .., y_t)^\top e(w'))}\nonumber
\end{equation}
where $f(\cdot)$ processes the conditioning context using a Transformer, and $e(\cdot)$
represents the weights of the softmax layer.

As depicted in in \figref{fig:arch},
the mixed character-subword Transformer consumes characters as input generates subwords as output. 
This figure only shows the decoder architecture,
since as the encoder that processes $\vx$ is a standard subword Transformer. 
Once a subword $w$ is emitted at time step $t$,
the characters of the subword $w$ are fed into the decoder for time steps $t+1$ to $t+|w|$,
and the next subword is generated at time step $t+|w|$, conditioned on all of the previously generated characters. 
 
 %

\subsection{Optimization}

The training objective for our latent segmentation translation model is 
$  \sum_{(\vx,\vy) \in \mathcal{D}} \log P_{\theta}(\vy|\vx)$ 
where $\mathcal{D}$ is the  training corpus consisting of parallel bilingual sentence pairs. 
Maximizing the training objective requires marginalization and the computation of the gradient of the log marginal likelihood.

\paragraph{Exact Marginalization.} Under our model, the probability of a subword only depends on the character-based
encoding of the conditioning context and not its segmentation, as in \eqref{eq:jointz2}.
This means that we can compute the log marginal likelihood for a single example $\vy$, exactly, using
the Dynamic Programming algorithm shown in Algorithm \ref{alg:dp}.
The core of the algorithm is line 3, where the probability of the prefix string $\vy_{0,k}$
is computed by summing terms corresponding to different segmentations.
Each term consists of the product of the probability of a  subword $\vy_{j,k}$ times the probability of its
conditioning context $(y_1, \ldots, y_j)$. The running time of the algorithm
is $\mathcal{O}(mT)$, where $T$ is the length of the string, and $m$ is the size of the longest subword unit in the vocabulary.   

\paragraph{Gradient Computation.}
We use automatic differentiation in PyTorch 
to backpropagate through the dynamic program in Algorithm \ref{alg:dp} and compute its gradient. 
Compared to a standard Transformer decoder, our mixed character-subword Transformer is 8x slower with a larger memory footprint, due to computation involved in the DP algorithm and large sequence length in characters. 
To address these issues, we reduce the number of transformer layers from 6 to 4, and accumulate 16 consecutive gradients before one update.

\subsection{Segmenting Target Sentences} 

\begin{algorithm*}[t] 
\caption{Dynamic Programming Encoding (DPE) for Subword Segmentation} 
\label{alg:best_seg} 
\begin{algorithmic} 
\REQUIRE $\vy$ is a sequence of $T$ characters, $V$ is a subword vocabulary, $m$ is the maximum subword length
\ENSURE Segmentation $\vz$ with highest posterior probability.
    \FOR{$k=1$ \TO $T$}
         \STATE { $\beta_k \leftarrow \max_{\left\{j \in [k-m,k-1] \,\mid\, \vy_{j,k} \in V\right\}}
          \beta_j + \log P_{\theta}(\vy_{j,k}|y_1,..,y_{j})$}
         \STATE { $b_k \leftarrow \mathrm{argmax}_{\left\{j \in [k-m,k-1] \,\mid\, \vy_{j,k} \in V\right\}} \beta_j + \log P_{\theta}(\vy_{j,k}|y_1,..,y_{j})$ }
            \ENDFOR
    \STATE{$\vz \leftarrow \textrm{backtrace}(b_1,..,b_T)$\hspace{5cm}  \COMMENT{backtrace the best segmentation using $\vb$}}
\end{algorithmic}
\end{algorithm*}


\label{sec:dpe-seg}

Once the mixed character-subword transformer is trained, it is used to segment the target side of a bilingual corpus.
We randomize the subword segmentation of source sentences using BPE dropout~\cite{provilkov2019bpe}.
Conditional on the source sentence, we use \Algref{alg:best_seg}, called Dynamic Programming Encoding (DPE) to
find a segmentation of the target sentence with highest posterior probability.
This algorithm is similar to the marginalization algorithm, where we use a max operation instead of log-sum-exp.
The mixed character-subword transformer is used only for tokenization, and a standard 
subword transformer is trained on the segmented sentences. For
inference using beam search, the mixed character-subword transformer is not needed.


\section{Experiments}

\begin{table}[t]
    \centering
    \begin{tabular}{@{}l@{~}l|cccc@{}}
            && \multicolumn{3}{c}{Number of sentences} & Vocab \\
            &&train & dev& test & size  \\
    \midrule  \midrule
       En-Hu &WMT09 & 0.6M & 2,051 & 2,525 & 32K\\
       En-De &WMT14 &4.2M & 3000 & 3003 & 32K\\
       En-Fi &WMT15 &1.7M & 1,500 & 1,370& 32K\\
       En-Ro &WMT16 & 0.6M & 1,999 & 1,999& 32K\\
       En-Et &WMT18 & 1.9M & 2,000 & 2,000& 32K\\
     \bottomrule
    \end{tabular}
    \caption{Statistics of the corpora.}
    \label{tab:data}
\end{table}

\paragraph{Dataset}

\begin{table*}[t!]
\begin{center}
    \begin{tabular}{l c c c cc c cc}
    \toprule
    \multicolumn{1}{c}{Method} &\hspace*{.3cm}& BPE &\hspace*{.3cm}& \multicolumn{1}{c}{BPE dropout} & \multirow{3}{*}{$\Delta_1$} &\hspace*{.3cm}& \multicolumn{1}{c}{This paper} & \multirow{3}{*}{$\Delta_2$}\\
    \cmidrule{1-1}
    \cmidrule{3-3}
    \cmidrule{5-5}
    \cmidrule{8-8}
    \multicolumn{1}{c}{Source segmentation} && BPE && BPE dropout &  && BPE dropout & \\
    \multicolumn{1}{c}{Target segmentation} && BPE && BPE dropout &  && DPE &\\
    \midrule 
    \midrule
      En$\rightarrow$De && 27.11 && 27.27 & +0.16 && 27.61 & +0.34\\
      En$\rightarrow$Ro && 27.90 && 28.07 & +0.17 && 28.66& +0.59\\
      En$\rightarrow$Et && 17.64 && 18.20 & +0.56 && 18.80 & +0.60\\
      En$\rightarrow$Fi && 15.88 && 16.18 & +0.30 && 16.89& +0.71\\   
      En$\rightarrow$Hu && 12.80 && 12.94 & +0.14 && 13.36& +0.42     \\ 
      \cmidrule{1-9}
      De$\rightarrow$En && 30.82 && 30.85 & +0.03 && 31.21 & +0.36\\
      Ro$\rightarrow$En && 31.67 && 32.56 & +0.89 && 32.99 & +0.43 \\
      Et$\rightarrow$En && 23.13 && 23.65 & +0.52 && 24.62 & +0.97\\
      Fi$\rightarrow$En && 19.10 && 19.34 & +0.24 && 19.87 & +0.53\\
      Hu$\rightarrow$En && 16.14 && 16.61 & +0.47 && 17.05 & +0.44\\       
    \midrule
      Average && 22.22 && 22.57 & +0.35 && 23.12 & +0.55\\
    \bottomrule
    \end{tabular}
    
\end{center}
\caption{Average test BLEU scores (averaged over $3$ independent runs)
  for $3$ segmentation algorithms, namely BPE~\citep{sennrich2016neural}, BPE
  dropout~\citep{provilkov2019bpe}, and our DPE algorithm on $10$ different
  WMT datasets.  $\Delta_1$ shows the improvement of BPE dropout compared
  to BPE, and $\Delta_2$ shows further improvement DPE compared to BPE dropout.
  All of the segmentation algorithms use the same subword dictionary with $32$K
  tokens shared between source and target languages. }
\label{tab:dropout}
\end{table*}

\begin{table*}[!t]
\small
    \centering
    \begin{tabular}{l}
      \toprule
      BPE source:\\
      Die G+le+is+anlage war so ausgestattet , dass dort \textcolor{blue}{elektr+isch} betrie+bene \textcolor{blue}{Wagen} eingesetzt\\werden konnten .\\
      DPE target:\\
      The railway system was equipped in such a way that \textcolor{blue}{electrical+ly} powered \textcolor{blue}{cart+s} could be used on it .\\
      BPE target:\\
      The railway system was equipped in such a way that \textcolor{blue}{elect+r+ically} powered \textcolor{blue}{car+ts} could be used on it .\\
         \midrule
      BPE source:\\
      \textcolor{blue}{Normalerweise} wird Kok+ain in kleineren Mengen und nicht durch \textcolor{blue}{Tunnel} geschm+ug+gelt .\\
      DPE target:\\
      \textcolor{blue}{Normal+ly} c+oca+ine is sm+ugg+led in smaller quantities and not through \textcolor{blue}{tunnel+s} .\\
      BPE target:\\
      \textcolor{blue}{Norm+ally} co+c+aine is sm+ugg+led in smaller quantities and not through \textcolor{blue}{tun+nels} .\\

      \bottomrule
    \end{tabular}
    \caption{Two examples of segmentation of English sentences given German inputs.}
    \label{tab:seg_src}
    \vspace{-3mm}
\end{table*}


We use {WMT09 for En-Hu}, WMT14 for En-De, WMT15 for En-Fi, WMT16 for En-Ro and WMT18 for En-Et.  
We utilize Moses toolkit\footnote{https://github.com/moses-smt/mosesdecoder} to pre-process all corpora, and preserve the true case of the text. Unlike \citet{lee2018deterministic}, we retain diacritics for En-Ro to retain the morphological richness.
We use all of the sentence pairs where the length of either side is less than 80 tokens for. training.
Byte pair encoding (BPE)~\citep{sennrich2016neural} is applied to all language pairs to construct a
subword vocabulary and provide a baseline segmentation algorithm. The statistics of all corpora is summarized in Table \ref{tab:data}.

\paragraph{Training with BPE Dropout.} We apply BPE dropout~\citep{provilkov2019bpe} to each mini-batch.
For each complete word, during the BPE merge operation, we randomly drop a particular merge with a probability of 0.05.
This value worked the best in our experiments.
A word can be split into different segmentations at the training stage, which helps improve the BPE baseline.

\paragraph{DPE Segmentation.} 
DPE can be used for target sentences, but its use for source sentences is not justified as source segmentations should
not be marginalized out. Accordingly, we use BPE dropout for segmenting source sentences.
That is, we train a mixed character-subword transformer to marginalize out the latent segmentations of a target sentence,
given a randomized segmentation of the source sentence by BPE dropout. 
After the mixed character-subword transformer is trained, it is used to segment the target sentences as describe in \secref{sec:dpe-seg}
for tokenization.

%

\begin{figure}[t]
\centering
\includegraphics[scale=0.35]{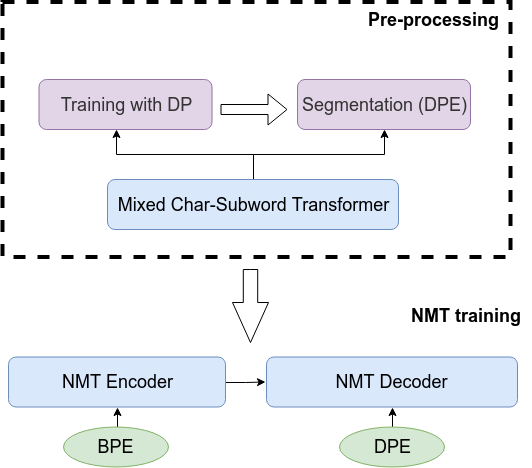}
\caption{The workflow of the proposed DPE approach.}
\label{fig:workflow}
\vspace{-.2cm}
\end{figure}

As summarized in \Figref{fig:workflow}, we first train a mixed character-subword transformer with dynamic programming.
Then, this model is frozen and used for DPE segmentation of target sentences.
Finally, a standard subword transformer is trained on 
source sentences segmented by BPE dropout and target sentences segmented by DPE.
The mixed character-subword transformer is not needed for translation inference.

\paragraph{Transformer Architectures.} We use transformer models to train three translation models on BPE, BPE dropout, and DPE corpora. 
We make use of \textit{transformer base} for all of the experiments.

\subsection{Main Results}

Table \ref{tab:dropout} shows the main results. 
%
%
First, we see that BPE dropout consistently outperforms BPE across language pairs.
In Table \ref{tab:dropout}, the column labeled to $\Delta_1$ shows the improvement of BPE dropout over BPE.
This gain can be attributed to the robustness of the NMT model to the segmentation error on the source side,
as our analysis in Section \ref{sec:analysis} will confirm. 
Second, we observe further gains resulted from DPE compared to BPE dropout.
The column labeled $\Delta_2$ shows the improvement of DPE over BPE dropout.
DPE provides an average improvement of 0.55 BLEU over BPE dropout and
BPE dropout provides an average improvement of 0.35 BLEU over BPE.
As our proposal uses BPE dropout for segmenting the source, 
we attribute our BLEU score improvements to a better segmentation of the target language with DPE.
Finally, compared to BPE for segmenting the source and target, our proposed segmentation method
results in large improvements in the translation quality, up to 1.49 BLEU score improvements
in Et$\rightarrow$En.  

\subsection{Segmentation Examples} 
\tabref{tab:seg_src} shows examples of target sentences segmented using DPE and BPE and the corresponding source sentences.  
In addition, \tabref{tab:seg_freq} presents the top 50 most common English words that result in a disagreement between BPE and DPE
segmentations based on the Et$\rightarrow$En corpus.
For DPE, for each word, we consider all segmentations produced and show the segmentation that attains the highest frequency of usage
in \tabref{tab:seg_freq}.
As can be observed, DPE produces more linguistically plausible morpheme-based subwords compared to BPE. 
For instance, BPE segments \textit{``carts"} into \textit{``car"+``ts"}, as both \textit{``car"} and \textit{``ts"} are common subwords
and listed in the BPE merge table. By contrast DPE segments \textit{``carts"} into \textit{``cart"+``s"}.
We attribute the linguistic characteristics of the DPE segments to the fact that DPE conditions the segmentation of a target word
on the source sentence and the previous tokens of the target sentence, as opposed to BPE, which mainly makes use of frequency of subwords,
without any context. 

DPE generally identifies and leverages some linguistic properties, \eg plural, antonym, normalization, verb tenses, etc.
However, BPE tends to deliver less linguistically plausible segmentations, possibly due to its greedy nature and the lack of
context. We believe this phenomenon needs further investigation, \ie the contribution of source vs. target context in DPE segmentations, and 
a quantitative evaluation of linguistic nature of word fragments produced by DPE.
We will leave this to future work.



\begin{table}[t!]
\small
    \centering
    \begin{tabular}{l l}
      \toprule
      BPE & DPE (ours) \\
      \midrule \midrule
      recognises & recognise~+~s \\
      advocates &	 advocate~+~s \\
      eurozone &  euro~+~zone\\
      underlines &  underline~+~s\\
      strengthens &  strengthen~+~s\\
      entrepreneurship &  entrepreneur~+~ship \\
      acknowledges &  acknowledge~+~s\\
      11.30 &  11~+~.30\\
      wines & wine~+~s\\
      pres~+~ently &present~+~ly \\
      f~+~illed &  fill~+~ed\\
      endors~+~ement& endorse~+~ment\\
      blo~+~c & bl~+~oc\\
      cru~+~cially&  crucial~+~ly \\
      eval~+~uations & evaluation~+~s\\
      tre~+~es&tr~+~ees\\
      tick~+~ets & tick~+~et~+~s\\
      predic~+~table&predict~+~able\\
      multilater~+~alism &multilateral~+~ism\\
      rat~+~ings & rating~+~s\\
      predic~+~ted&  predict~+~ed\\
      mo~+~tives&  motiv~+~es\\
      reinfor~+~ces & reinforce~+~s\\
      pro~+~tocols & protocol~+~s\\
      pro~+~gressively & progressive~+~ly \\
      sk~+~ill& ski~+~ll\\
      preva~+~ils & prevail~+~s\\
      decent~+~ralisation& decent~+~ral~+~isation\\
      sto~+~red& stor~+~ed\\
      influ~+~enz~+~a & influen~+~za\\
      margin~+~alised & marginal~+~ised\\
      12.00 & 12~+~.00\\
      sta~+~ying& stay~+~ing\\
      intens~+~ity& intensi~+~ty\\
      rec~+~ast & re~+~cast\\
      guid~+~eline & guide~+~line\\
      emb~+~arked &embark~+~ed \\
      out~+~lines& outline~+~s\\
      scen~+~ari~+~os& scenario~+~s\\
      n~+~ative& na~+~tive\\
      preven~+~tative & prevent~+~ative\\
      hom~+~eland& home~+~land\\
      bat~+~hing& bath~+~ing\\
      endang~+~ered& endanger~+~ed \\
      cont~+~inen~+~tal & continent~+~al\\
      t~+~enth& ten~+~th\\
      vul~+~n~+~era~+~bility & vul~+~ner~+~ability\\
      realis~+~ing & real~+~ising\\
      t~+~ighter&tight~+~er \\
     \bottomrule
    \end{tabular}
    \caption{Word fragments obtained by BPE vs. DPE.
    The most frequent words that resulted in a disagreement between
    BPE and DPE segmentations on $Et \to En$ are shown.}
    \label{tab:seg_freq}
    \vspace{-3mm}
\end{table}

\subsection{Analysis}
\label{sec:analysis}

\paragraph{Conditional Subword Segmentation.} 
One of our hypothesis for the effectiveness of subword segmentation with DPE is that it conditions the segmentation of the target on the source language. 
To verify this hypothesis, we train mixed character-subword Transformer solely on the target language sentences in the bilingual training corpus using the \emph{language model} training objective. 
This is in contrast to the mixed character-subword model used in the DPE segmentation of the main results in Table \ref{tab:dropout}, where the model is conditioned on the source language and trained on the sentence pairs using a \emph{conditional language model} training objective. 
Once the mixed character-subword Transformer language model is trained, it is then used to segment the target sentence of the bilingual corpus in the pre-processing step before a translation model is trained. 
\begin{table}[t]
    \centering
    \begin{tabular}{lccc}
     \toprule
    Source  & BPE drop & BPE drop  & BPE drop  \\
    Target  & BPE drop & LM DPE &  DPE  \\
    \midrule \midrule   
       En$\rightarrow$Ro & 28.07  & 28.07 & 28.66\\
       En$\rightarrow$Hu & 12.94 & 12.87 &   13.36\\
       \midrule
       Ro$\rightarrow$En & 32.56 & 32.57 &   32.99\\
       Hu$\rightarrow$En &16.61& 16.41 &    17.05\\
      \bottomrule
    \end{tabular}
    \caption{DPE-LM learns a segmentation of the target based on language modelling, which is \emph{not} conditioned on the source language. }
    \label{tab:lm-share}
    \vspace{-3mm}
\end{table}

Table \ref{tab:lm-share} shows the results. It compares the unconditional language model (LM) DPE vs the conditional DPE for segmenting the target language, where we use BPE dropout for segmenting the source language.  
We observe that without the information from the source, LM DPE is on-par to BPE, and is significantly outperformed by conditional DPE. This observation confirms our hypothesis that segmentation in NMT  should be source-dependent.

\begin{figure}[t]
\centering
\includegraphics[scale=0.4]{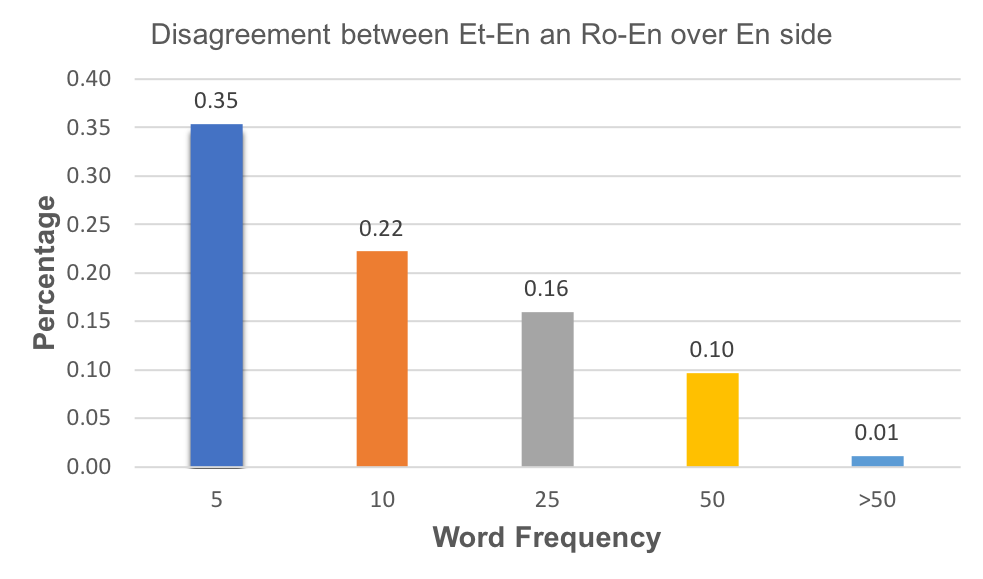}
\caption{Disagreement of DPE segments between Et-En and Ro-En over English vocabulary}
\label{fig:tgt:seg}
\end{figure}

We are further interested in analyzing the differences of  the target language segmentation depending on the source language. 
For this analysis, we filtered out a multilingual parallel corpus from WMT, which contains parallel sentences in three languages English, Estonian and Romanian.
That is, for each English sentence we have the corresponding sentences in Et and Ro.
We then trained two DPE segmentation models for the translation tasks of Et$\rightarrow$En and Ro$\rightarrow$En, where English is the target language.
Figure \ref{fig:tgt:seg} shows when conditioning the segmentation of the target on different source languages, DPE can lead to different segmentations even for an identical multi-parallel corpus. The differences are more significant for low frequency words.

Another aspect of DPE segmentation method is its dependency on the \emph{segmentation} of the source. 
As mentioned, we segment the target sentence \emph{on the fly} using our mixed character-subword model \emph{given} a randomized segmentation of the source produced by BPE dropout. 
That means during the training of the NMT model where we use BPE dropout for the source sentence, the corresponding target sentence may get a different DPE segmentation given the randomized segmentation of the source sentence. 
We are interested in the effectiveness of the target segmentation if we commit to a fixed DPE segmentation conditioned on the BPE segmentation of the input. 
Table \ref{tab:fly} shows the results. We observe that there is a marginal drop when using the fixed DPE, which indicates that the encoder can benefit from a stochastic segmentation, while the decoder prefers a deterministic segmentation corresponding to the  segmentation of the source. 

\begin{table}[t]
    \centering
    \begin{tabular}{lccc}
     \toprule
    Source  & BPE drop  & BPE drop  \\
    Target  & DPE Fixed & DPE On The Fly  \\
    \midrule \midrule   
       En$\rightarrow$Ro &  28.58 & 28.66\\
       En$\rightarrow$Hu &  13.14 & 13.36\\
       En$\rightarrow$Et &  18.51 &  18.80\\
       \midrule
       Ro$\rightarrow$En &  32.73 &   32.99\\
       Hu$\rightarrow$En & 16.82 &    17.05\\
       Et$\rightarrow$En & 24.37 &    24.62\\
      \bottomrule
    \end{tabular}
    \caption{``DPE Fixed'' obtains a fixed  segmentation of the target sentence given the BPE-segmented source sentence, whereas ``DPE On The Fly'' obtain the best segmentation of the target sentence given a randomized segmentation of the source produced by BPE dropout. }
    \label{tab:fly}
\vspace{-3mm}    
\end{table}

\begin{table}[h]
\begin{center}
    \begin{tabular}{lccc}
    \toprule
    Source  & BPE drop & BPE drop          & BPE drop \\
    Target  & BPE      & BPE drop          & DPE \\
    \midrule \midrule
      En$\rightarrow$Ro & 28.04 & 28.07 & 28.66\\
      En$\rightarrow$Et & 18.09 & 18.20 & 18.80\\
      \midrule
      Ro$\rightarrow$En &  32.40  & 32.56 & 32.99\\
      Et$\rightarrow$En &  23.52  & 23.65  & 24.62\\
    \bottomrule
    \end{tabular}
\end{center}    
\caption{BLEU score of different target segmentation methods. 
}
\label{tab:tgt}
\vspace{-3mm}
\end{table}


\paragraph{DPE vs BPE.} 
We are interested to compare the effectiveness of DPE versus BPE 
for the target, 
given BPE dropout as  the same  segmentation method for the source. 
Table \ref{tab:tgt} shows the results. As observed, target segmentation with DPE consistently outperforms BPE, leading to up to .9 BLEU score improvements. 
We further note that using BPE dropout on the target has a similar performance to BPE, and it is consistently outperformed by DPE. 
%

\begin{figure}[h]
\centering
\includegraphics[scale=0.4]{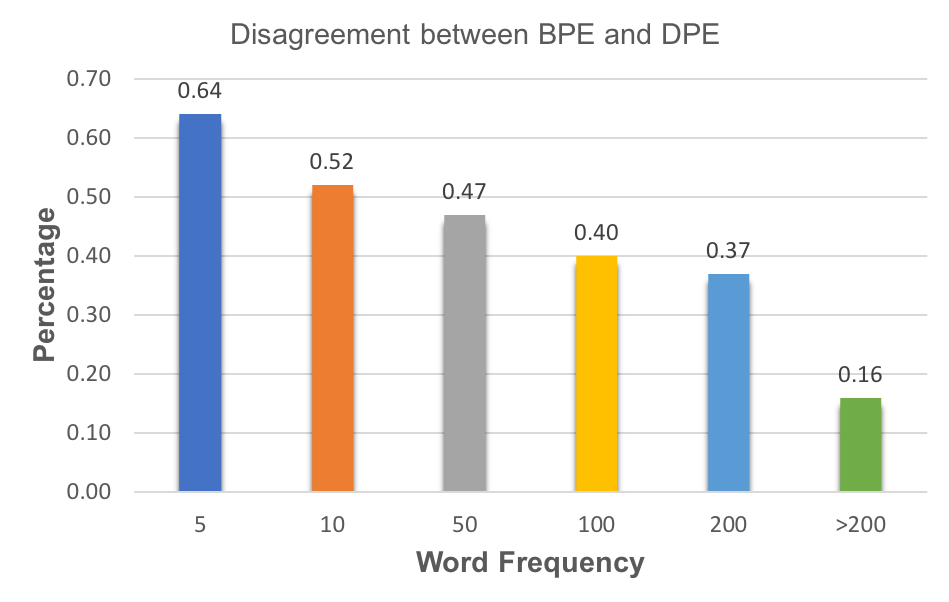} 
\caption{Disagreement of segments between BPE and DPE over Estonian vocabulary.}
\label{fig:completion}
\end{figure}

We further analyze the segmentations produced by DPE vs BPE. 
Figure \ref{fig:completion} shows the percentage of the target words which have  different segmentation with BPE and DPE, for different word frequency bands in En$\rightarrow$Et translation task. 
We observe that for Estonian words whose occurrence is up to 5 in the training set, the disagreement rate between DPE and BPE is 64\%. The disagreement rate decreases as we go to words in higher frequency bands. 
This may imply that the main difference between the relatively large BLEU score difference between BPE and DPE is due to their different segmentation mainly for low-frequency words. 

\begin{figure}[h!]
\centering
        \includegraphics[width=8cm]{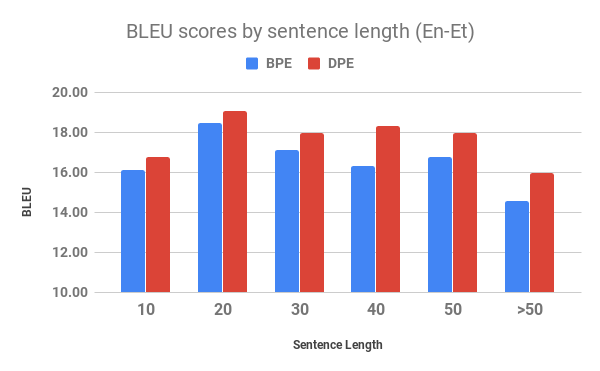}
\caption{BLEU scores of BPE vs DPE by the lengths of sentences for En$\rightarrow$Et. }
\label{fig:bleu_len}
\end{figure}

We further plot the distribution of BLEU scores by the length of target sentences. As shown in \figref{fig:bleu_len}, DPE demonstrates much better gains on the longer sentences, compared with the BPE version.

\section{Conclusion}

This paper introduces \textit{Dynamic Programming Encoding} 
in order to incorporate the information of the source language into subword segmentation of the target language. 
Our approach utilizes dynamic programming for marginalizing the latent segmentations when training, and inferring the highest probability  segmentation when tokenizing. 
Our comprehensive experiments show impressive improvements 
compared to state-of-the-art segmentation methods in NMT, \ie BPE and its stochastic variant BPE dropout.


\section*{Acknowledgment}
We would like to thank the anonymous reviewers, Taku Kudo, Colin Cherry and George Foster for their comments and suggestions on this work.
The computational resources of this work are supported by the Google Cloud Platform (GCP), and by the Multi-modal Australian ScienceS Imaging and Visualisation Environment (MASSIVE) (\url{www.massive.org.au}).
This material is partly based on research
sponsored by Air Force Research Laboratory and DARPA
under agreement number FA8750-19-2-0501. The U.S.
Government is authorized to reproduce and distribute
reprints for Governmental purposes notwithstanding any
copyright notation thereon.

\bibliography{acl2020}
\bibliographystyle{acl_natbib}

\appendix


\end{document}